\pdfoutput=1

\documentclass[11pt]{article}
\usepackage{authblk}

\usepackage{acl}

\usepackage[T1]{fontenc}
\usepackage[utf8]{inputenc}

\usepackage{times}

\usepackage{microtype}

\usepackage{graphicx}

\usepackage{xspace}

\usepackage{url}

\usepackage{placeins}

\usepackage{siunitx}  

\usepackage{adjustbox}

\usepackage{enumitem}

\usepackage[colorinlistoftodos,prependcaption]{todonotes}

\presetkeys%
    {todonotes}%
    {inline,backgroundcolor=yellow}{}
    
\usepackage{booktabs}

\usepackage[title]{appendix}

\usepackage{caption}

\usepackage{acro}
\usepackage{hyperref}

\usepackage[capitalise]{cleveref}

\newcommand{\hids}[1]{\textsc{\lowercase{#1}}\xspace} 

\crefformat{chapter}{\S#2#1#3}
\crefmultiformat{chapter}{\S\S#2#1#3}{ and~#2#1#3}{, #2#1#3}{, and~#2#1#3}

\crefformat{section}{\S#2#1#3}
\crefmultiformat{section}{\S\S#2#1#3}{ and~#2#1#3}{, #2#1#3}{, and~#2#1#3}

\DeclareAcronym{nlp}{short = NLP, long = Natural Language Processing}

\newif\iffinal\finalfalse

\usepackage{acro}

\usepackage{ascii}
\usepackage{newunicodechar}

\usepackage[normalem]{ulem}

\title{Lessons Learned from a Citizen Science\\ Project for Natural Language Processing}

\author[1]{Jan-Christoph Klie}
\author[1]{Ji-Ung Lee}
\author[1,2]{Kevin Stowe}
\author[1,3]{Gözde Gül Şahin}
\author[1,4]{\authorcr Nafise Sadat Moosavi}
\author[1]{Luke Bates}
\author[1]{Dominic Petrak}
\author[1]{\authorcr Richard Eckart de Castilho}
\author[1]{Iryna Gurevych}
 
\affil[1]{Ubiquitous Knowledge Processing Lab (UKP Lab)\protect\\ Department of Computer Science and Hessian Center for AI (hessian.AI)\protect\\ Technical University of Darmstadt \protect\\ \url{www.ukp.tu-darmstadt.de}\protect\vspace{.6em}}
\affil[2]{Educational Testing Service}
\affil[3]{KUIS AI, Koç University}
\affil[4]{Department of Computer Science, The University of Sheffield}

\date{}

\begin{document}
\maketitle

\begin{abstract}
    Many Natural Language Processing (NLP) systems use annotated corpora for training and evaluation.
However, labeled data is often costly to obtain and scaling annotation projects is difficult, which is why annotation tasks are often outsourced to paid crowdworkers.
Citizen Science is an alternative to crowdsourcing that is relatively unexplored  in the context of NLP.
To investigate whether and how well Citizen Science can be applied in this setting, we conduct an exploratory study into engaging different groups of volunteers in Citizen Science for NLP by re-annotating parts of a pre-existing crowdsourced dataset.
Our results show that this can yield high-quality annotations and attract motivated volunteers, but also requires considering factors such as scalability, participation over time, and legal and ethical issues.
We summarize lessons learned in the form of guidelines and provide our code and data to aid future work on Citizen Science.\footnote{\url{https://github.com/UKPLab/eacl2023-citizen-science-lessons-learned}}

\end{abstract}

\section{Introduction}
\label{sec:intro}

Data labeling or \emph{annotation} is often a difficult, time-consuming, and therefore expensive task. 
Annotations are typically drawn from domain experts or are crowdsourced. 
While experts can produce high-quality annotated data, they are expensive and do not scale well due to their relatively low number~\citep{sorokinUtilityDataAnnotation2008}. 
In contrast, crowdsourcing can be relatively cheap, fast, and scalable, but is potentially less suited for more complicated annotation tasks~\citep{drutsaCrowdsourcingPracticeEfficient2020}. 
Another approach is using Citizen Science, which describes the participation and collaboration of volunteers from the general public with researchers to conduct science~\citep{haklayWhatCitizenScience2021}.
Over the past decade, Citizen Science platforms, which rely on unpaid volunteers to solve scientific problems, have been used for a wide variety of tasks requiring human annotation~\citep{handPeoplePowerNetworks2010}, e.g., classifying images of galaxies~\citep{lintottGalaxyZooMorphologies2008} or for weather observation~\citep{leeperObservationalPerspectivesClimate2015}.

\begin{figure*}[t!]
    \centering
    \includegraphics[width=.99\textwidth]{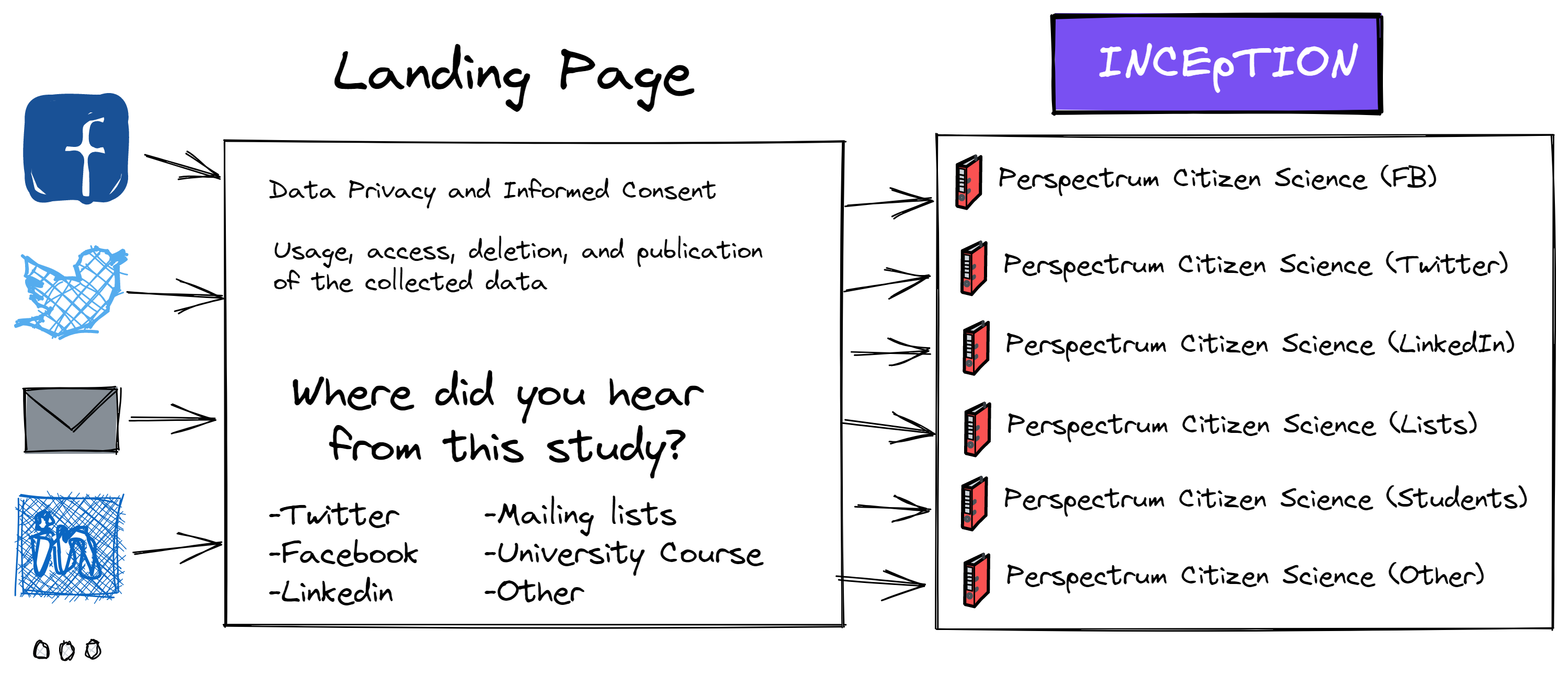}
    \caption{We advertised our project via various social media, mailing lists and university courses. Volunteers then are onboarded via the landing page and donated annotations via INCEpTION.}
    \label{fig:annotation_pipeline}
\end{figure*}

While Citizen Science has been shown to produce high-quality annotations in ecological or environmental projects~\citep{kosmalaAssessingDataQuality2016}, its potential has so far not been investigated in depth for \ac{nlp}.
Our goal in this work is to assess the practicality of undertaking annotation campaigns for \ac{nlp} via Citizen Science.
We analyze  whether volunteers actually react to our calls and participate, how the resulting quality is compared to crowdsourcing, what the benefits and shortcomings are and what needs to be taken into account when conducting such a project.
We especially are interested in differences between annotators recruited via different channels, which we investigate by advertising to different social media platforms, \ac{nlp}-related mailing lists, and university courses.
To explore this possibility, we use the \hids{Perspectrum} dataset~\citep[][CC-BY-SA]{chenSeeingThingsDifferent2019} that focuses on the task of stance detection and can be motivated by fighting misinformation and promoting accurate debate in internet discussions.
We replicated a portion of the annotations in this dataset using citizen scientists instead of crowdworkers. 
To accomplish this goal, we designed an annotation workflow that is suitable for Citizen Science and allows us to recruit volunteers across a variety of platforms.

Our contributions are the following:

\begin{enumerate}
    \item We provide a systematic study on Citizen Science across different channels and analyze turnout and quality. For this, we re-annotate parts of the  \hids{Perspectrum} dataset using Citizen Science and compare these to the original, crowdsourced annotations. 
    \item We provide guidelines and recommendations on how to successfully conduct a Citizen Science project for \ac{nlp} annotation and discuss critical legal and ethical aspects.
    \item We provide a platform for future Citizen Science projects that handles onboarding, anonymous access, work assignment and the annotating itself.
\end{enumerate}

Our results show that using Citizen Science for linguistic annotation can result in high-quality annotations, but that attracting and motivating people is critical for its success, especially in the long-term.
We were able to attract 98 volunteers when conducting our Citizen Science project which resulted in 1,481 annotations over 2 months, thereby re-annotating around 10\% of the original dataset.
We find that annotations obtained through mailing lists and university students were of high quality when comparing them to the original, adjudicated crowdsourced data.
We thus conclude that Citizen Science projects have the potential to be applied to \ac{nlp} annotation if they are conceptualized well, but are best suited for creating smaller datasets.

\section{Background}
\label{sec:background}
Prior work has developed various means and strategies for annotating large datasets.
So far, annotation studies in NLP mainly use domain-experts or crowdworkers, or a mix of both~\citep{nguyenCombiningCrowdExpert2015}.
Crowdsourcing in particular has received increasing attention over the past decade~\citep{wangPerspectivesCrowdsourcingAnnotations2013}.

\paragraph{Paid Experts}
Recruiting domain experts (e.g., linguists) for annotation studies has been a widely accepted method to generate linguistically annotated corpora.
Famous examples are the Brown Corpus~\citep{francisBrownCorpusManual1979} or the Penn Treebank~\citep{marcusBuildingLargeAnnotated1993}.
While the resulting datasets are of the highest quality, domain experts are often few, and such annotation studies tend to be slow and expensive~\citep{sorokinUtilityDataAnnotation2008}.
Although many researchers moved on to annotation studies that recruit crowdworkers, expert annotations are still necessary in various fields, e.g., biomedical annotations~\citep{hobbsECOCollecTFCorpusAnnotated2021}.

\paragraph{Crowdsourcing} 
To accelerate the annotation process and reduce costs, researchers have utilized crowdsourcing as a means to annotate large corpora~\citep{snowCheapFastIt2008}.
The main idea behind crowdsourcing is that annotation tasks that do not require expert knowledge can be assigned to a large group of paid non-expert annotators.
This is commonly done via crowdsourcing platforms such as Amazon Mechanical Turk (AMT) or Upwork and has been successfully used to annotate various datasets across different tasks and domains~\citep{derczynskiBroadTwitterCorpus2016, habernalArgumentationMiningUserGenerated2017}.
Previous work compared the quality between crowdsourcing and expert annotations, showing that many tasks can be given to crowdworkers without major impact on the quality of annotation~\citep{snowCheapFastIt2008, hovyExperimentsCrowdsourcedReannotation2014, dekuthyFocusAnnotationTaskbased2016}.

Although crowdworkers can substantially accelerate annotation, crowdsourcing requires careful task design and is not always guaranteed to result in high quality data \citep{danielQualityControlCrowdsourcing2018}.
Moreover, as annotators are compensated not by the time they spend but rather by the number of annotated instances, they are compelled to work fast to maximize their monetary gain---which can negatively affect annotation quality~\citep{drutsaCrowdsourcingPracticeEfficient2020} or even result in spamming~\citep{hovyLearningWhomTrust2013}.
It can also be difficult to find crowdworkers for the task at hand, for instance due to small worker pools for languages other than English~\citep{pavlickLanguageDemographicsAmazon2014, frommherzCrowdsourcingEcologicallyValidDialogue2021} or because the task requires special qualifications~\citep{tauchmannInfluenceInputData2020}.
Finally, the deployment of crowdsourcing remains ethically questionable due to undervalued payment~\citep{fortAmazonMechanicalTurk2011,cohenEthicalIssuesCorpus2016}, privacy breaches, or even psychological harm on crowdworkers~\citep{shmueliFairPayEthical2021}.

\paragraph{Games with a Purpose} A related but different way to collect annotations from volunteers is \textit{games with a purpose}, i.e., devising a game in which participants annotate data~\citep{chamberlainAddressingResourceBottleneck2008, venhuizenGamificationWordSense2013}.
Works propose games for different purposes and languages.
For instance, anaphora annotation (PhraseDetectives, \citealt{poesioPhraseDetectivesUtilizing2013}), dependency syntax annotation (Zombilingo, \citealt{fortCreatingZombilingoGame2014}), or collecting idioms~\citep{eryigitGamifiedCrowdsourcingIdiom2022}.  
It has been shown that if a task lends itself to being gamified, then it can attract a wide audience of participants and can be used to create large-scale datasets~\citep{vonahnGamesPurpose2006}.
Finally, \citet{lydingApplicabilityCombiningImplicit2022} investigate games with a purpose in the context of (second) language learning to simultaneously crowdsource annotaions from learners as well as teachers.
One such example is Substituto, a turn-based, teacher-moderated game for learning verb-particle constructions~\citep{aranetaSubstitutoSynchronousEducational2020}.   
We do not consider gamification in this work, as enriching tasks with game-like elements requires considerable effort and cannot be applied to every task.

\paragraph{Citizen Science}

Citizen Science broadly describes participation and collaboration of the general public (the citizens) with researchers to conduct science~\citep{haklayWhatCitizenScience2021}.
Citizen Science is a popular alternative approach for dataset collection efforts, and has been successfully applied in cases of weather observation~\citep{leeperObservationalPerspectivesClimate2015}, counting butterflies~\citep{holmesOntarioButterflyAtlas1991} or birds~\citep{national2020christmas}, classifying images of galaxies~\citep{lintottGalaxyZooMorphologies2008} or monitoring water quality~\citep{addy2010volunteer}.
Newly-emerging technologies and platforms further allow researchers to conduct increasingly innovative  Citizen Science projects, such as the prediction of influenza-like outbreaks~\citep{leeFluWatchersEvaluationCrowdsourced2021} or the classification of animals from the Serengeti National Park~\citep{swansonSnapshotSerengetiHighfrequency2015}.
\textit{LanguageARC} is a Citizen Science platform for developing language resources~\citep{fiumaraLanguageARCDevelopingLanguage2020}.
It is however not open yet to the public to create projects and does not easily allow conducting a Citizen Science meta-study as we do in this work.
One work using LanguageARC is by \citet{fortUseCitizenScience2022} (LD) who collected resources to evaluate bias in language models. 
They did not investigate the impact of using different recruitment channels which we do.
Other projects using LanguageARC are still running and it is too early to derive recommendations from.

\begin{figure*}[!t]
    \centering
    \includegraphics[width=.99\textwidth]{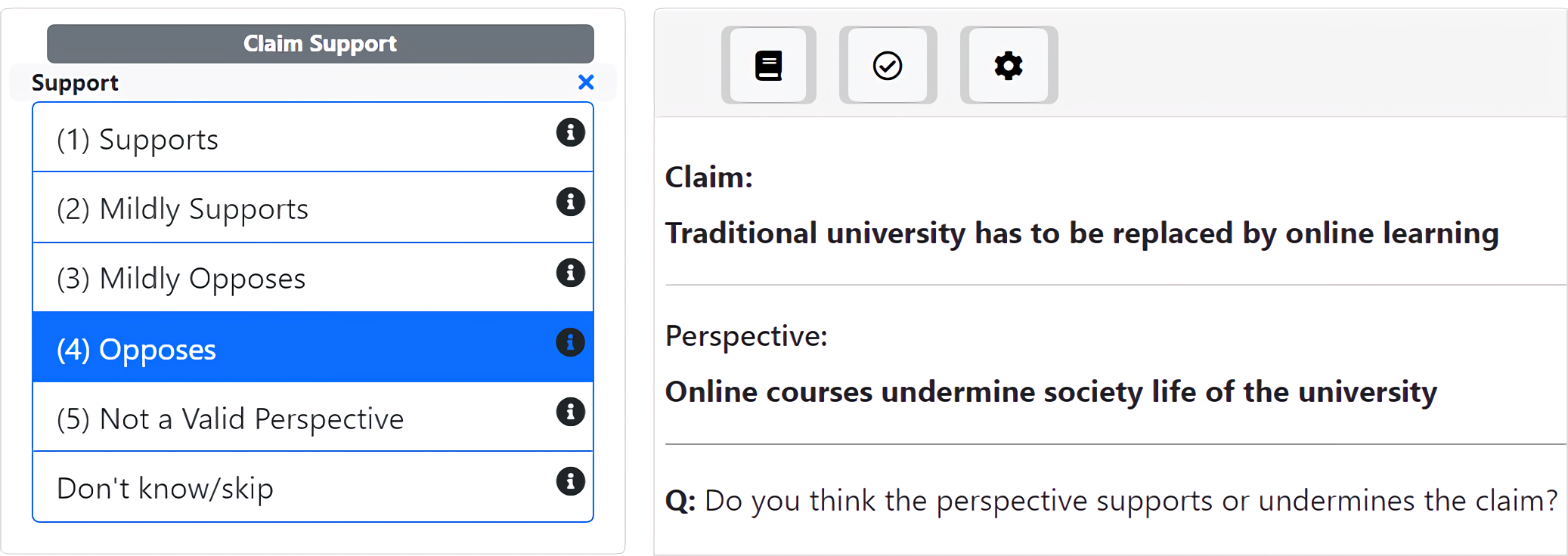}
    \caption{Assigning a label to an instance in the INCEpTION text annotation platform.}
    \label{fig:annotation_interface}
\end{figure*}%
Compared to crowdsourcing, Citizen Science participants are volunteers that do not work for monetary gain.
Instead, they are often motivated intrinsically. 
For instance, they may have a personal interest on positively impacting the environment~\citep{westVariationsMotivationsEnvironmental2021}, or in altruism~\citep{rotmanDynamicChangesMotivation2012}.
Asking for unpaid work also entails various issues like finding good ways of how to attract volunteers, and ethical considerations~\citep{resnikFrameworkAddressingEthical2015, rasmussenCitizenScienceEthics2019} that need to be addressed (cf. \cref{sec:takeaways}).
Intrinsic motivation also has the potential of resulting in higher-quality annotations compared to crowdsourcing. 
For instance, \citet{leeAnnotationCurriculaImplicitly2022} find in their evaluation study with citizen scientists that their participants may have been willing to take more time annotating for the sake of higher annotation accuracy.
However, as their main goal was to conduct an evaluation study for their specific setup, this finding cannot be generalized to other Citizen Science scenarios.
So far, only \citet{tsuengCitizenScienceMining2016} provide a direct comparison between crowdsourcing and Citizen Science and show that volunteers can achieve similar performance in mining medical entities in scientific texts.
They recruit participants through different channels such as newspapers, Twitter, etc., but do not compute channel-specific performance, making it difficult to assess whether the quality of the resulting annotation depends on the recurrent channel.
In contrast, in the present work, we explicitly consider the recruitment channel in our evaluation and furthermore provide a discussion and guidelines for future Citizen Science practitioners.
Also, it attracts intrinsically (not only fiscally)  motivated volunteers that are often skilled in the task and can provide high-quality annotations, thus potentially combining the advantages of expert annotations and crowdsourcing.
Relying on unpaid annotators entails several issues, including attracting volunteers and ethical considerations~\citep{resnikFrameworkAddressingEthical2015, rasmussenCitizenScienceEthics2019} that need to be taken into account~(see \cref{sec:takeaways}).

\section{Study Design}

\begin{figure*}[t]
    \includegraphics[width=\textwidth]{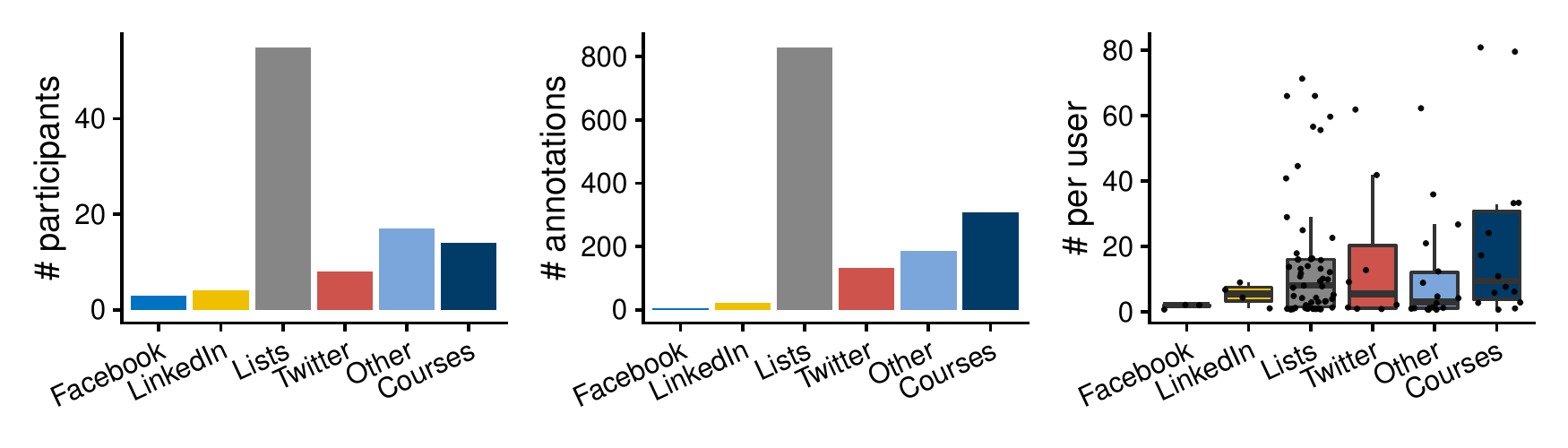}
    \caption{Participants, annotations and annotations grouped by the channel via they were recruited. It can be seen that overall, most participants and annotations were contributed by annotators recruited via mailing lists. Annotators from mailing lists and courses yielded the volunteers who contributed the most individually.}
    \label{fig:participation}
\end{figure*}

To study the feasibility of Citizen Science for \ac{nlp} annotation, we asked volunteers recruited via various channels to re-annotate an existing, crowdsourced dataset.
The general setup is described in \cref{fig:annotation_pipeline}.
To conduct a systematic study, we identified the following four necessary steps:
1) Identifying a suitable dataset~(\cref{sec:dataset_collection});
2) Selecting suitable recruitment channels to advertise our project on~(\cref{sec:outreach});
3) Building a landing page for onboarding participants that asks for informed consent and the channel from which they originated~(\cref{sec:landing_website});
4) Setting up the annotation editor to which participants are forwarded after the onboarding~(\cref{sec:inception_annotation}).

\subsection{Dataset selection}
\label{sec:dataset_collection}

We first conducted a literature review of relevant crowdsourced NLP datasets to identify the ones that could be accurately reproduced via Citizen Science. We assessed datasets for the following two criteria: 
1) \textbf{Availability}: the dataset must be publicly available to make proper comparisons in terms of annotator agreement;
2) \textbf{Reproducibility}: the annotation setup including annotation guidelines needs to be reproducible to ensure similar conditions between citizen scientists and crowdworkers. 
We focused on datasets that are targeted towards contributing to social good to encourage volunteers to participate. 
Unfortunately, many inspected datasets did not fulfill both of these requirements. 
Overall, we identified two main issues while screening over 20 candidate datasets. 
First, many datasets used Tweets which impacted reproducibility as Twitter only allows researchers to publish the tweet identifiers.
This leads to irrecoverable instances when tweets were deleted.
Second was the lack of precise guidelines. 
For instance, many considered datasets about societal biases lack explicit descriptions of what is considered a stereotype.
As such biases are often also impacted by the respective cultural background of annotators, they are difficult to reproduce without specific guidelines. 

In the end, we decided on the stance detection task of the \hids{Perspectrum} dataset~\citep{chenSeeingThingsDifferent2019}.
The task provides clear instructions, publicly available data, and is motivated by social good (fighting misinformation/promoting accurate debate in internet discussions).
Each instance consists of a claim--perspective pair~(cf.~\cref{fig:annotation_interface}) and annotators are asked if the claim \textit{supports}, \textit{opposes}, \textit{mildly-supports}, \textit{mildly-opposes}, or is \textit{not a valid perspective}.
Following the original work, we also evaluated the annotations on a coarser tagset that only contains the categories for \textit{support}, \textit{oppose} and \textit{not a valid perspective}.
Overall, the dataset consists of $907$ claims and $8,370$ different perspectives which yield $11,805$ annotated instances.
In preliminary studies, we received further feedback that forcing annotators to provide an explicit label for each instance could lead to increasing frustration, especially for ambiguous or complicated instances.
To lessen the burden for our voluntary annotators and keep them motivated in the annotation task, we allowed them to skip instances (\textit{Don't know/skip}) which was not present in the original annotation editor for \hids{Perspectrum}.

\subsection{Recruitment channels}
\label{sec:outreach}
To recruit annotators, we advertised our project on three social media platforms, namely, Twitter, LinkedIn and Facebook.
Unfortunately, after creating the Facebook organization and advertising the project, the account was banned due to ``violating their community standards'' and has so far remained banned. 
One of our team members then promoted our annotation study on their personal Facebook to attract participation from this social media platform.
In addition, the team members advertised the work on Twitter and in relevant LinkedIn groups such as \textsc{Computational Linguistics} and \textsc{Machine Learning and Data Science}.

We further promoted the study via two external mailing lists (i.e., \textsc{corpora-list}, \textsc{ml-news}).
Late in the project, we received interest from other faculty to advertise the task in their courses---an offer that we gladly accepted.
For this, participation was completely voluntary and anonymous, students' grades were not affected by participation, and authors were not among the instructors.
To evaluate different recruitment channels separately, we asked participants on the landing page to answer the question: ``Where did you hear from this study?''. 
We also allowed volunteers to not disclose how they found out about the study, this is referred to as ``Other'' or ``Undisclosed'' in this paper.
Final participation counts are given in \cref{fig:participation}.
We deliberately limited our outreach, e.g. we did not use university social media accounts or colleagues with large follower bases.
Also, we made sure to not exhaust channels by posting too many calls for participation.

\subsection{Landing page}
\label{sec:landing_website}

We implemented a customizable landing page web application catering to the needs of Citizen Science projects.
The link to such a landing page was shared via the respective recruitment channels.
The landing page contained information about the study itself, its purpose, its organizers, which data we collected, and its intended use.
This landing page toolbox is designed so that it can easily be adapted to future Citizen Science projects.
To allow project creators to use an annotation editor of their choice, we designed the toolbox to act as an intermediary that collects a participant's consent for the actual annotation study.
This ensures that only participants that have been properly informed and have explicitly provided their consent are given access to the study.
For future Citizen Science projects, the tool further assists organizers through the landing page creation process to foster an ethical collection of data by asking several questions, that are listed in the appendix.

\subsection{Annotation editor}
\label{sec:inception_annotation}

INCEpTION~\citep{klieINCEpTIONPlatformMachineAssisted2018} offers a configurable, web-based platform for annotating text documents at span,  relation and document levels.
To make it usable in Citizen Science scenarios, we extended the platform with three features, namely, (1) the ability to join a project through a link, (2) support for anonymous guest annotators, and (3) a dynamic workload manager.
Allowing citizen scientists to participate in the project anonymously as guests without any sign-up process substantially reduced the entry barrier and made it easier for us to satisfy data protection policies.
The same is true for the ability of joining a project through an invite link.
Upon opening the link, annotators were greeted with the annotation guidelines and were directly able to start annotating.
Finally, we implemented a dynamic workload manager that takes as input the desired number of annotators per document and then automatically forwards annotators directly to the document instances requiring annotation.
Upon finishing annotating an instance, INCEpTION was configured to automatically load and display the next instance for annotation, similar to popular crowdsourcing platforms.
We also included rules for handling other issues that may occur with voluntary annotations such as recovering instances that annotators have started to work on but then abandoned.
Additionally, we modified the existing user interface to improve the annotation workflow.
This mainly included implementing a dedicated labeling interface that allows users to select a single label for an instance via a radio button group.
Annotation of an instance thus required two user actions: first, selecting the document label, and second, confirming the annotation, thereby moving on to the next document.

\section{Results}
\label{sec:results}

\begin{figure*}[!t]
\noindent\begin{minipage}{.49\textwidth}
\centering
\includegraphics[width=.99\textwidth]{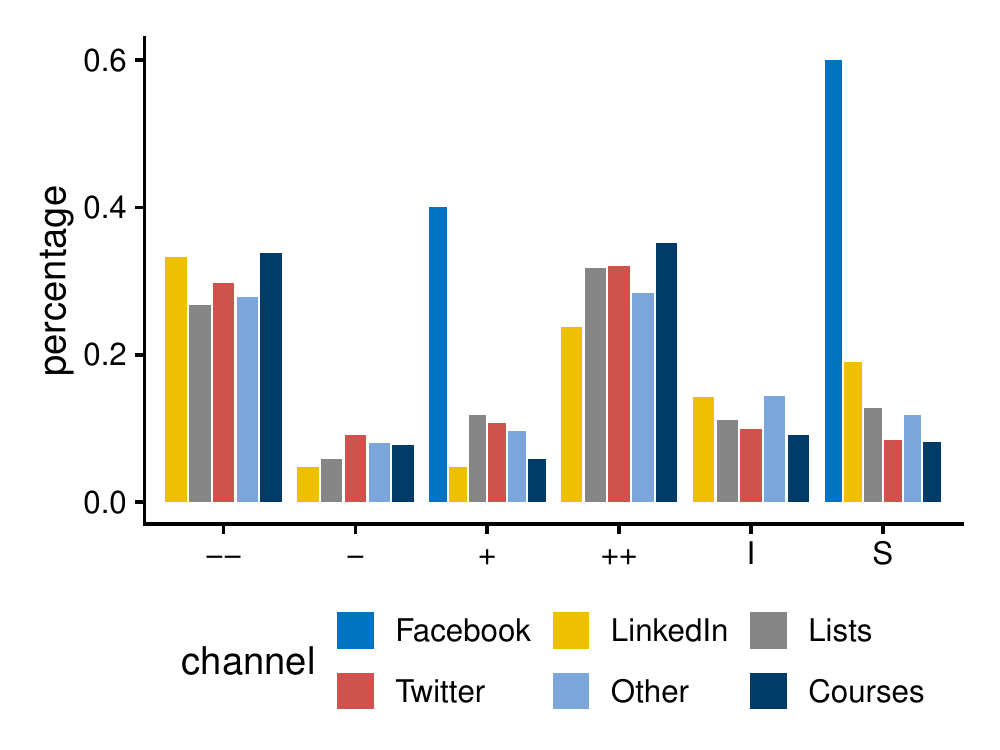}
\captionof{figure}{Label distribution grouped by channel. Labels are \textit{supports} (++), \textit{mildly-supports} (+), \textit{mildly-opposes} (-), \textit{opposes} (-{}-), \textit{not a valid perspective} (I) and \textit{Skip} (S).}
\label{fig:label_distribution}            
\end{minipage}%
\hfill
\begin{minipage}{.49\textwidth}
\centering
\includegraphics[width=.99\textwidth]{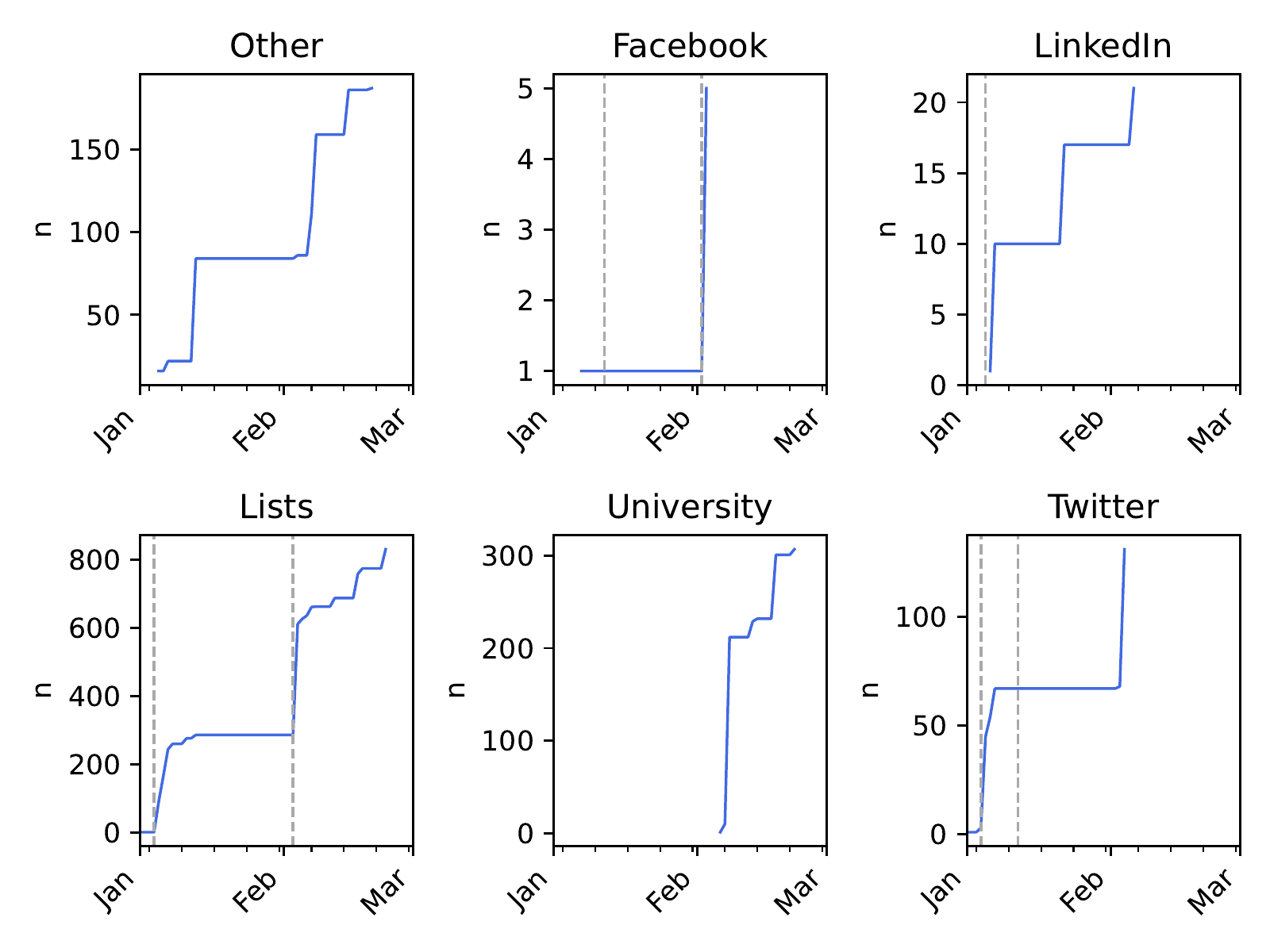}
\captionof{figure}{Annotations made over time. Vertical lines indicate when calls on the respective channels have been posted.}
\label{fig:participation_over_time}            
\end{minipage}
\end{figure*}

We conducted our study between January and March 2022 and promoted the task in successive rounds across all recruitment channels.
In total, we were able to recruit 98 participants who provided 1481 annotations resulting in 906 fully annotated instances. 
Each instance with at least one annotation has received on average 1.63 annotations. 
Detailed statistics are provided in the appendix.

\paragraph{Participation}
To identify promising channels for future Citizen Science studies, we report the number of annotators per channel, the total number of annotations per channel and per user (cf.~\cref{fig:participation}).
Overall, we find that the most effective channel for public outreach are mailing lists (55 participants). 
Asking students in university courses to participate was the second most effective with 14 participants.
Facebook, LinkedIn, and Twitter only yielded three, four, and eight participants respectively.
We further find a highly skewed distribution of annotations per user, as many annotators only provide a few annotations while a few annotators provide many annotations.
For instance, the most active annotators were two students who provided $\sim$80 annotations as well as six participants from mailing lists who provided $\sim$60--80 annotations each.
For Twitter and ``undisclosed'', only a single annotator made over 60 annotations.
We also find that on average, participants from university courses provided the most annotations per person.
When looking at participation over time (see \cref{fig:participation_over_time}), we observe increased activity in annotations made after the call for participation has been posted to the respective channel.
For many channels, the count quickly flattens. 
Interestingly, Twitter sees a second spike long after the post was made.
We attribute it to people sharing the post in our community quite a while after the initial release.
We did not track whether individual volunteers came back for another round of annotations after their initial participation.

\paragraph{Coverage}

Overall, our 98 volunteers have provided 1,481 annotations to 906 unique instances (approximately 8\% of the original dataset) over two months.
This is comparable to other Citizen Science projects like~\citet{fortUseCitizenScience2022}, which had 102 participants in total.
They annotated three tasks and collected 2347, 2904 and 220 submissions over eight months.
Table~\ref{tab:annotation_coverage} shows the resulting coverage of our Citizen Science annotation study. 
While this still leaves room for improvement, the number of annotations collected nonetheless shows that Citizen Science can be viable in real life settings and is a promising direction to investigate in further studies, especially for creating focused and smaller-scale resources.

\begin{table}[bh]
    \centering
    \caption{Claims, claim clusters, and individual claim-perspective pairs that have been annotated at least once.
    We call the set of a claim and a perspective together with its paraphrases a claim cluster.}
    \begin{tabular}{@{}lrrS@{}}
\toprule
 Name                  &   \# Annotated &   \# Total &   {\% Annotated} \\
\midrule
 Claims                &           388 &       907 &      42.7784  \\
 Clusters              &           739 &      5092 &      14.513   \\
 Total                 &           906 &     11805 &       7.67471 \\
\bottomrule
\end{tabular}
    \label{tab:annotation_coverage}
    \vspace{-3mm}
\end{table}

\paragraph{Quality}
In terms of annotation quality, we find that most channels yield annotations that highly agree with the gold labels (cf.~\cref{tab:annotation_quality}), even though our annotations are not adjudicated yet.
We further find that volunteers from university courses and mailing list show the highest accuracy, followed by Twitter and ``undisclosed''.
Only LinkedIn yields lower accuracy than 70\% on the coarse label set.

For the majority of channels (with the exception of Facebook and LinkedIn), we only see a skip percentage of $\sim$10\% (cf.~\cref{fig:label_distribution}).
This indicates our volunteers are actually willing to spend time and effort to solve the task at hand, as adding a ``Don't know/skip'' option in crowdsourcing usually is an invitation for workers to speed through the tasks and not provide useful annotations.
The exception is Facebook, where we find that a majority of the annotations from Facebook were labeled as \textit{I don't know/skip} (3 out of 5).
Further analysis of the label distribution grouped by channel (cf.~\cref{fig:label_distribution}) shows that all channels except for Facebook display a similar distribution in terms of annotated labels.
This indicates that we can expect a rather stable annotation performance across citizen scientists recruited from different channels.

\begin{table}[ht]
    \centering
    \caption{Annotation accuracy compared to the crowdsourced and adjudicated data from \hids{Perspectrum}. The five annotations from Facebook (three of them were skipped) and \textit{Don't know/skip} annotations are omitted. }
    \begin{tabular}{@{}lSS@{}}
\toprule
 Channel   &   {Coarse} &   {Fine} \\
\midrule
 University  &   0.919492 & 0.822034 \\
 LinkedIn  &   0.692308 & 0.615385 \\
 Mailing Lists        &   0.898876 & 0.816479 \\
 Undisclosed   &   0.841667 & 0.75     \\
 Twitter   &   0.849462 & 0.731183 \\
\bottomrule
\end{tabular}
    \label{tab:annotation_quality}
    \vspace{-4mm}
\end{table}

\section{Discussion and Takeaways}
\label{sec:takeaways}
Here we present lessons learned, discuss legal challenges and ethical considerations, as well as provide guidelines for future Citizen Science projects.

\paragraph{Channel-dependent differences}
Our results clearly differ across recruiting channels.
We find that overall, Facebook and LinkedIn have the lowest turnout and accuracy when compared to the gold labels, followed by Twitter.
Our assumption for the overall low participation is that our network for these channels was not large enough.
Advertising our study to NLP-related and university-internal mailing lists and university courses yielded the highest number of participants who also provided the most and best-quality annotations. 
Although our results show that students may outperform participants from other channels, we also acknowledge that this may not always be a viable option to recruit citizen scientists.
Overall, our findings indicate that it is important to address the respective target groups that may be interested in a specific study.
However, we also note that continuously advertising Citizen Science studies to the same channels may have a negative impact, as it can cause participation fatigue and lead to fewer volunteers participating.
One possible solution could be the use of LanguageARC~\citep{fiumaraLanguageARCDevelopingLanguage2020} from the LDC and centralize calls for participation.

\paragraph{Motivating volunteers}
In contrast to crowdsourcing, there is no monetary or other extrinsic motivation that could be used to attract Citizen Science annotators.
Thus, annotator motivation is a crucial question for Citizen Science studies.
As \cref{fig:participation_over_time} shows, citizen scientists can be quickly motivated to participate, but can also quickly lose interest in a given annotation study.
This can become an issue with a low number of participants, yet our results also indicate that we were able to find highly-motivated participants  (8 out of 98  in our results).

Compared to other groups, university students in particular provided a high amount of quality annotations.
Considering the findings by \citet{phillipsAssessingCitizenScience2018}, who do not find statistical differences in terms of quality between students participating for course credit vs. no extrinsic reward---asking students to participate in such projects as part of their coursework might be another good option, but needs to ensure an ethical data collection. 
For instance, such an approach has been used to annotate the Georgetown University Multilayer Corpus~\citep{zeldesGUMCorpusCreating2017}.
Nonetheless, one remaining question is how to keep participants motivated and participate in several sessions as our results indicate that a vast majority of our volunteers only participated in a single session and that participation quickly stops shortly after a call has been posted to the respective channel.

Finally, we want to emphasize the inclusion of a \textit{Don't know/skip} option for Citizen Science annotators. 
Whereas in crowdsourcing studies, annotators may exploit such an option to increase their gain~\citep{hovyLearningWhomTrust2013}, from the feedback we got during our pilot study, it is crucial to keep volunteers motivated for Citizen Science.
For this work, we did not provide a survey that asks about the motivation, as we thought that this might deter potential participants
We however suggest that future studies provide such a survey that is as unintrusive as possible to further analyze why participants take part in the respective annotation project.

\paragraph{Legal challenges}
One substantial challenge in implementing Citizen Science studies is the potentially wide outreach they can have and, consequently, the varying kinds of data protection regulations they have to oblige. 
To preempt any potential issues that can arise---especially when data that can be used to identify a person (personal data, e.g. obtained during a survey or login credentials) is involved---we recommend researchers who plan to implement a Citizen Science study consider the most strict regulations that are widely accepted.

For the GDPR~\citep{GDPR}, currently one of the strictest data protection regulations, we recommend researchers to explicitly ask voluntary participants for their informed consent when collecting personal information.
This includes informing participants beforehand about (1) the purpose of the data collection, (2) the kind of personal and non-personal data collected, (3) the planned use of the data, (4) any planned anonymization processes for publication, and finally, (5) how participants can request access, change, and deletion of the data.
We further recommend assigning one specific contact person for any questions and requests for access, change, or deletion of the data. 
This may seem like additional work when compared to crowdsourcing, but transparent and open communication is one of the key factors to build trust---which is necessary for voluntary participants to consider such studies and provide high-quality annotations.
Finally, participants should be informed and agree to the  annotations donated being published under a permissive license.

\paragraph{Ethical and economical considerations}
Although Citizen Science can substantially reduce annotation costs, we emphasize the importance of considering an ethical deployment that does not compromise the trust of the participants.
Moreover, given increasing concerns regarding the ownership and use of collected data~\citep{arrieta-ibarraShouldWeTreat2018}, one should grant participants full rights to access, change, delete, and share their own personal data~\citep{jonesNonrivalryEconomicsData2020}. 
This ensures that participants are not exploited for ``free labor''---in contrast to approaches like reCAPTCHA~\citep{vonahnReCAPTCHAHumanBasedCharacter2008}, where humans are asked to solve a task in order to gain access to services.
Whereas CAPTCHAs were initially intended to block malicious bots, they are becoming increasingly problematic due to their deployment and use by monopolizing companies which raises ethical concerns~\citep{avanesiNotRobotAm2022}.
It is especially important to take the data itself into consideration; exposing volunteers to toxic, hateful, or otherwise sensitive speech should be avoided if they are not informed about it beforehand.

\paragraph{Recommendations}

Overall, we derive the following recommendations for future Citizen Science studies.
1) our call for annotations resonated the most with the target group that is likely to benefit the most from contributing to it: \ac{nlp} researchers coming from mailing lists and university students.
Therefore, the target audience should be carefully selected, for instance by identifying topic-specific mailing lists or respective university courses.
This further means that the purpose of data collection should be made clear and that the results should be made publicly available.
2) the research question of the study should conform to the respective ethical and legal guidelines of the potential target group which should clearly be communicated to make the project accountable.
3) participation should be easy with clearly formulated annotation guidelines and, moreover, the annotation itself should be thoroughly tested beforehand to ensure that participants do not get frustrated due to design errors or choices.
For instance, in our preliminary study, we got the feedback that some instances are frustrating to annotate and hence added an option to skip.
4) analyzing participation over time shows that a Citizen Science project has to be continuously advertised in order to stay relevant and achieve high participation. Otherwise, it will be forgotten quickly.
This can be done by sharing status updates or creating preliminary results.
Fifth, we recommend asking about user motivation before, during or after the annotation with a survey to better understand the participants and their demographics.

\section{Conclusion}
\label{sec:conclusion}

In this work, we presented an exploratory annotation study for utilizing Citizen Science for \ac{nlp} annotation.
We developed an onboarding process that can easily be adapted to similar projects and evaluated Citizen Science annotations for re-annotating an existing dataset.
Furthermore, we extended the INCEpTION platform, a well-known open-source semantic annotation platform, with a dynamic workload manager and functionality for granting access to external users without registration. 
This enables its usage for Citizen Science projects. 
We advertised the study via Twitter, Facebook, LinkedIn, mailing lists, and university courses and found that participants from mailing lists and university courses are especially capable of providing high-quality annotations.
We further discuss legal and ethical challenges that need to be addressed when conducting Citizen Science projects and provide general guidelines for conducting future projects that we would like to have known before starting. 
Overall, we conclude that Citizen Science can be a viable and affordable alternative to crowdsourcing, but is limited by successfully keeping annotators motivated.
We will make our code and data publicly available to foster more research on Citizen Science for \ac{nlp} and other disciplines.

\paragraph{Future Work}

We see the following directions for further research and evaluation to better understand in which settings Citizen Science can be applicable and how to use it best.
Here, we used \textsc{perspectrum} as the dataset to annotate and mentioned in the participation calls that it benefits the social good. 
Therefore, it would be interesting to conduct more projects and see which datasets are suitable as well as whether volunteers participate, even if there is no extrinsic motivation.
Then, it can also be tested how annotator retention develops, especially when project are running longer. The call for participation itself could also be investigated for the impact it has on turnout, motivation and quality.

\section{Limitations}

Throughout this article, we analyzed whether Citizen Science applies to linguistic annotation and showed that we can attract volunteers that donate a sizeable number of high-quality annotations.
This work, however, comes with limitations that should be taken into account and tackled in future work.
First, we based our analysis on a single annotation campaign and dataset that we advertised as being relevant for the social good. 
Therefore we suggest conducting more such annotation projects, also with different kinds of tasks.
Second, we did not perform a user survey that for instance asked for user motivation.
This is why we can only speculate about the motivation of our participants and suggest future works to explicitly prepare such a survey.
Third, using Facebook as a channel might be viable, but we were not able to properly analyze it, as our account was blocked shortly after creation and never was reinstantiated.
Finally, based on participation and annotation numbers, we see Citizen Science as more of an option for annotating smaller datasets, or longer-term projects that are more actively advertised than in our study which took place over two months and for which we deliberately limited the outreach.

\section*{Acknowledgments}

We thank our anonymous reviewers, Michael Bugert and Max Glockner for their detailed and helpful comments to improve this manuscript.
We are especially grateful for the discussions with Michael and Anne-Kathrin Bugert regarding our study setup.
Finally, we thank the many volunteers that donated annotations for this project, as this work would not have been possible without their generous participation.

This work has been funded by the German Research Foundation (DFG) as part of the Evidence project (GU 798/27-1), UKP-SQuARE (GU 798/29-1), INCEpTION (GU 798/21-1) and PEER (GU 798/28-1), and within the project “The Third Wave of AI” funded by the Hessian Ministry of Higher Education, Research, Science and the Arts (HWMK). Further, it has been funded by the German Federal Ministry of Education and Research and HMWK within their joint support of the National Research Center for Applied Cybersecurity ATHENE.

\bibliography{bibliography, bibliography_jck}

\begingroup
\onecolumn
\begin{appendices}

\section{Landing Page}
\label{sec:appendix_landing_page}

\begin{center}
\includegraphics[height=.95\textheight]{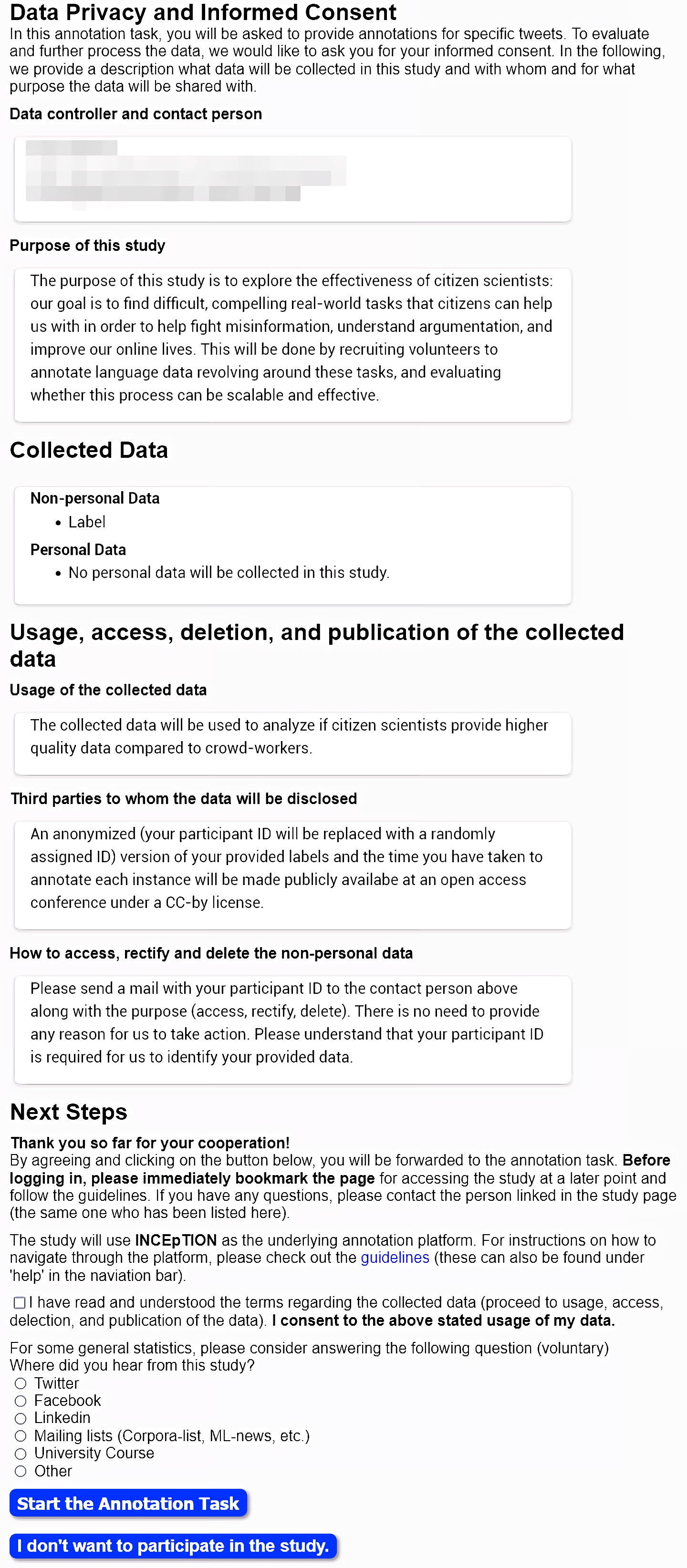}  
\end{center}

\section{Annotation Guidelines}
\label{sec:appendix_annotation_guidelines}

\begin{center}
\includegraphics[height=.95\textheight]{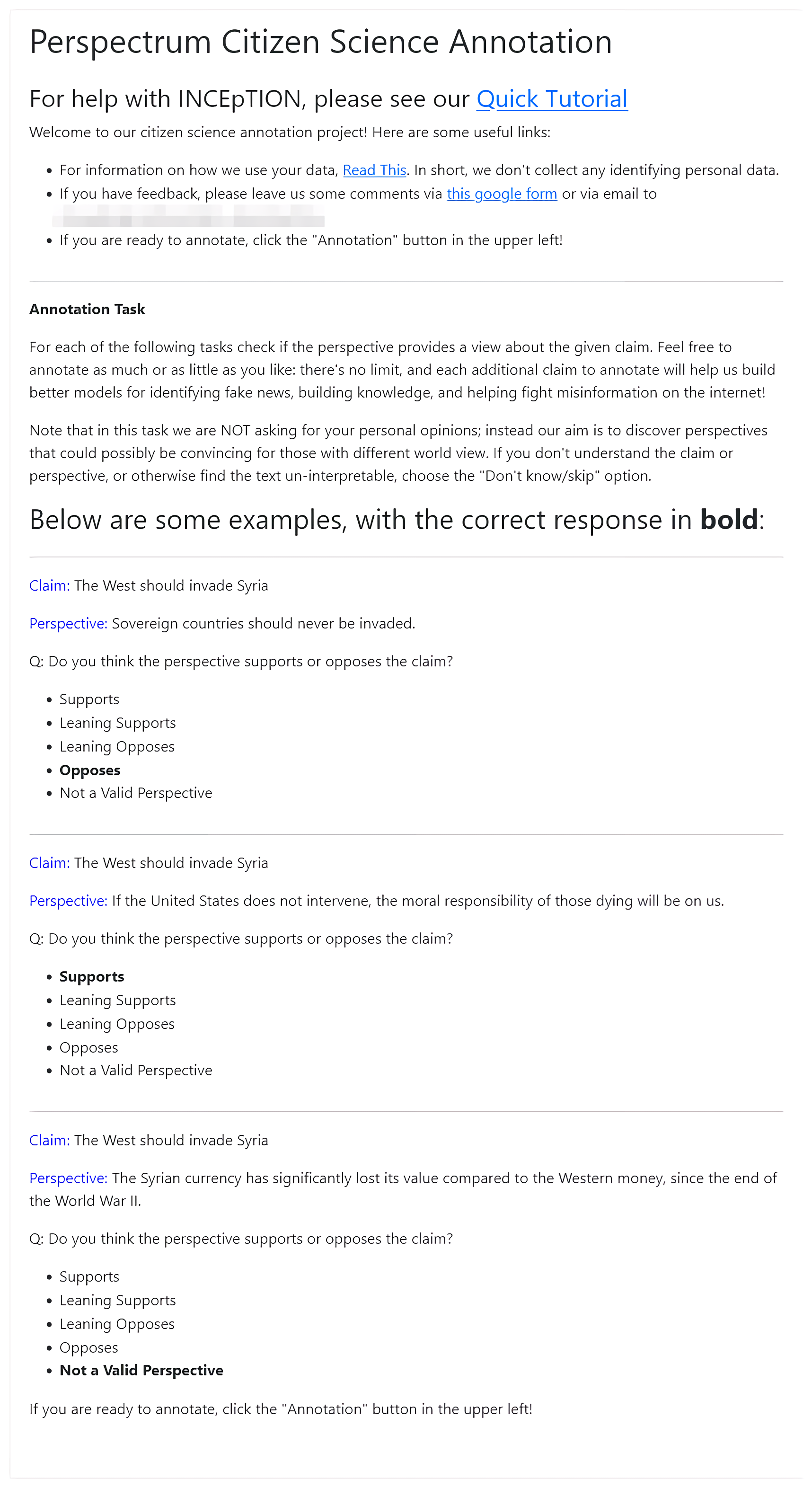}  
\end{center}

\section{Questions to keep in mind for a citizen science project}
\label{sec:appendix_questions_future}

\begin{itemize}[topsep=0pt,itemsep=-1ex,partopsep=1ex,parsep=1ex]
    \item What is the purpose of the study?
    \item What kind of personal and non-personal data will be collected?\footnote{We provided some pre-defined suggestions such as \textit{Name} or \textit{IP} for personal data and \textit{Label} for non-personal data with the possibility to add more in our landingpage module.}
    \item If there is a questionnaire involved, what questions will it involve?
    \item How will the data be used?
    \item Is a publication of the data planned and if so, which data will be published and will it be anonymized?
    \item How can participants request access, change, or deletion of their data?
\end{itemize}%

\section{Project Statistics}
\label{sec:appendix_numbers}

\subsection{Number of participants}

In addition to the plots visualizing the number of participants (c.f.~\cref{fig:participation}), we also list the raw numbers in \cref{tab:appendix_num_participants}.

\begin{table}[ht]
    \centering
    \begin{tabular}{@{}lr@{}}
\toprule
Channel  & Participants  \\
\midrule
Courses & 14 \\
Facebook & 3  \\
LinkedIn & 4  \\
Lists       & 55 \\
Twitter  & 8  \\
Undisclosed  & 17  \\
\bottomrule
\end{tabular}
    \caption{Number of participants per channel.}
    \label{tab:appendix_num_participants}
\end{table}

\subsection{Annotation statistics}

In addition to the plots visualizing the annotation counts and label distribution (c.f.~\cref{fig:label_distribution}), we also list the raw numbers in \cref{tab:appendix_label_distribution}.

\begin{table}[ht]
    \centering
    \caption{Label distribution grouped by channel. Labels are \textit{supports} (++), \textit{mildly-supports} (+), \textit{mildly-opposes} (-), \textit{opposes} (-{}-), \textit{not a valid perspective} (I) and \textit{Skip} (S).}      \begin{adjustbox}{max width=\textwidth}

\begin{tabular}{@{}lrrrrrrrSSSSSS@{}}
    \toprule
    Channel  & Total & \multicolumn{6}{c}{Counts}     & \multicolumn{6}{c}{Percentage}                            \\ 
    \cmidrule(l{3pt}r{3pt}){1-1}\cmidrule(l{3pt}r{3pt}){2-2}\cmidrule(l{3pt}r{3pt}){3-8}\cmidrule(l{3pt}r{3pt}){9-14}
                    & &  + &  ++ &  - & -{}- &  I &   S & {+}     & {++}    & {-}     & {-{}-}  & {I}     & {S}     \\ \midrule
        Courses     & 307 & 18 & 108 & 24 &  104 & 28 &  25 & 5.86319 & 35.1792 & 7.81759 & 33.8762 & 9.12052 & 8.14332 \\
        Facebook    & 5 &  2 &   0 &  0 &    0 &  0 &   3 & 40      & 0       & 0       & 0       & 0       & 60      \\
        LinkedIn    & 21 &  1 &   5 &  1 &    7 &  3 &   4 & 4.7619  & 23.8095 & 4.7619  & 33.3333 & 14.2857 & 19.0476 \\
        Lists       & 830 & 98 & 264 & 48 &  222 & 92 & 106 & 11.8072 & 31.8072 & 5.78313 & 26.747  & 11.0843 & 12.7711 \\
        Twitter     & 131 & 14 &  42 & 12 &   39 & 13 &  11 & 10.687  & 32.0611 & 9.16031 & 29.771  & 9.92366 & 8.39695 \\
        Undisclosed & 187 & 18 &  53 & 15 &   52 & 27 &  22 & 9.62567 & 28.3422 & 8.02139 & 27.8075 & 14.4385 & 11.7647 \\ \bottomrule
    \end{tabular}
\end{adjustbox}

    \label{tab:appendix_label_distribution}
\end{table}

\end{appendices}
\endgroup

\end{document}